\documentclass[runningheads]{llncs}
\usepackage{graphicx}

\usepackage{tikz}
\usepackage{comment}
\usepackage{amsmath,amssymb} 
\usepackage{color}
\usepackage{orcidlink}
\usepackage{multirow}
\newcommand*\samethanks[1][\value{footnote}]{\footnotemark[#1]}

\usepackage[accsupp]{axessibility}  

\begin{document}
\pagestyle{headings}
\mainmatter
\def\ECCVSubNumber{2596}  

\title{Robust Landmark-based Stent Tracking in X-ray Fluoroscopy} 

\titlerunning{Robust Landmark-based Stent Tracking in X-ray Fluoroscopy}
%
\author{Luojie Huang\inst{1}\orcidlink{0000-0002-4314-8959}\thanks{Work was done during an internship at United Imaging Intelligence America.}\and
Yikang Liu\inst{2}\orcidlink{0000-0003-1069-1215} 
\and Li Chen\inst{3}\samethanks 
\and Eric Z. Chen\index{Chen,Eric Z.} \inst{2}
\and Xiao Chen\inst{2}
\and Shanhui Sun\inst{2} \thanks{Corresponding author.}
}
\authorrunning{L. Huang et al.}
%
\institute{Johns Hopkins University, Baltimore, MD, USA \and
United Imaging Intelligence, Cambridge, MA, USA \email{shanhui.sun@uii-ai.com} \and University of Washington, Seattle, WA, USA
}


\maketitle

\begin{abstract}
In clinical procedures of angioplasty (i.e., open clogged coronary arteries), devices such as balloons and stents need to be placed and expanded in arteries under the guidance of X-ray fluoroscopy. Due to the limitation of X-ray dose, the resulting images are often noisy. To check the correct placement of these devices, typically multiple motion-compensated frames are averaged to enhance the view. Therefore, device tracking is a necessary procedure for this purpose. Even though angioplasty devices are designed to have radiopaque markers for the ease of tracking, current methods struggle to deliver satisfactory results due to the small marker size and complex scenes in angioplasty. In this paper, we propose an end-to-end deep learning framework for single stent tracking, which consists of three hierarchical modules: a U-Net for landmark detection, a ResNet for stent proposal and feature extraction, and a graph convolutional neural network for stent tracking that temporally aggregates both spatial information and appearance features. The experiments show that our method performs significantly better in detection compared with the state-of-the-art point-based tracking models. In addition, its fast inference speed satisfies clinical requirements. 

\end{abstract}

\section{Introduction}
\label{sec:introduction}

\begin{figure}[!t]
    \centering
    \includegraphics[width=\textwidth]{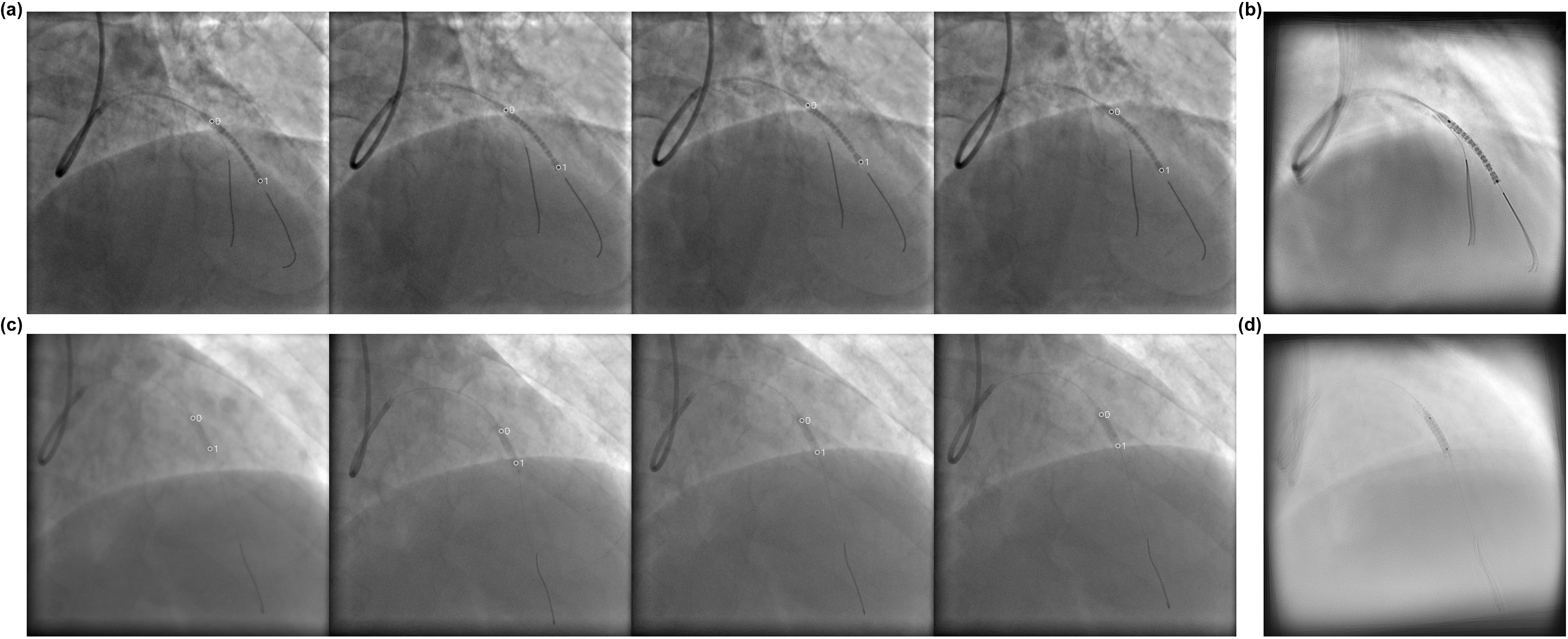}
    \caption{Examples of stent tracking with the proposed method (a,c) and stent enhancement based on the tracking results (b,d). a and c show four frames from a video.}
    \label{fig:teaser}
\end{figure}

Coronary artery disease (CAD) is one of the primary causes of death in most developed countries~\cite{CAD}. The current state-of-the-art treatment option for blocked coronary arteries is the percutaneous coronary intervention (PCI) (Fig.~\ref{fig:teaser}). During this minimally invasive procedure, a catheter with a tiny balloon (the tracked dark object in Fig.~\ref{fig:teaser}c) at the tip is put into a blood vessel and guided to the blocked coronary artery. Once the catheter arrives at the right place, the balloon is inflated to push the artery open, restoring room for blood flow. In most cases, a stent, which is a tiny tube of wire mesh (Fig.~\ref{fig:teaser}), is also placed in the blocked artery after the procedure to support the artery walls and prevent them from re-narrowing. Intraoperative X-ray fluoroscopy is commonly used to check the location of stent/balloon before expansion. However, stent visibility is often limited (Fig.~\ref{fig:teaser}a and c) under X-ray because the minimal level radiation dose out of safety concerns. Furthermore, stents keep moving rapidly with heartbeat and breathing in the complicated environment of patients' anatomy.

Compared to other physical approaches, such as invasive imaging or increasing radiation dose, a more cost-effective solution is to enhance the stent appearance through image processing (a.k.a., digital stent enhancement), as shown in Fig.~\ref{fig:teaser}b and d. A common method is to track the stent motion, separate the stent layer from the background layer, and average the stent layers from multiple frames after motion compensation. Stent tracking is achieved by tracking two radiopaque balloon markers that locate at two ends of the stent (Fig.~\ref{fig:teaser}). 

Stent tracking for enhancement remains quite challenging due to multiple reasons. First, the balloon markers are very small compared to the whole field of view (FOV) while the movements are large. Second, the scenes in PCI procedures are very complex: organs and other devices can form a noisy background and 3D organs and devices can be projected into different 2D images from different angles. Third, stent enhancement requires high localization accuracy and low false positives. Fourth, fast tracking speed is needed to meet the clinical requirements (e.g., 15fps for a 512x512 video). Fifth, data annotations are limited just like other medical imaging applications.

Current stent tracking methods lie under the tracking-by-detection category and assume only one stent presents in the FOV. They first detect all possible radiopaque balloon marker from each frame, and then identify the target stent track based on motion smoothness~\cite{elastic,compreh} or consistency score~\cite{lu,wang}. However, these methods are prone to large detection and tracking errors caused by strong false alarms. 
Deep learning techniques dramatically improve detection and tracking accuracy. However, it is difficult to apply these techniques to the stent tracking problem because of the the small object size and complex PCI scenes and overfitting issue on small dataset. Therefore, we tackle the above issues by incorporating some basic prior knowledge into our framework design. For example, the stent has distinctive hierarchical features: two dark markers that can be detected by low-level features and complicated patterns between the marker pairs, such as wire, mesh and balloon tubes, to be recognized from high-level semantic analysis. Additionally, the association of marker pairs in different frames requires long temporal dimension reasoning to tolerate inaccurate detections in certain frames with limited image quality. Moreover, most deep learning frameworks for keypoint tracking problems (such as human pose tracking) train detection and tracking modules separately. However, it is generally harder to detect small markers from a single X-ray image than keypoints from natural images due to the limited object features and complex background. 

Therefore, we propose an end-to-end trainable deep learning framework that consists of three hierarchical modules: a landmark detection module, which is a U-Net~\cite{unet} trained with small patches to detect potential balloon markers with local features; a stent proposal and feature extraction module, which is a ResNet~\cite{resnetv2} trained with larger stent patches to extract high-level features located between detected marker pairs; and a stent tracking module, which is a graph convolutional neural network (GCN) to associate marker pairs across frames using the combination of extracted features and spatial information. Our ablation study demonstrated that end-to-end learning of the whole framework can greatly benefit the performance of final stent tracking. For example, detector can be boosted by incorporating the feedback from trackers by learning to suppress some false positives that cause bad outcome in trackers. 

In summary, the major contributions of this paper are as follows: 1) We propose the first deep learning method, to our best knowledge, to address the single stent tracking problem in PCI X-ray fluoroscopy. 2) To handle the challenge of tracking small stents in complex video background, we propose an end-to-end hierarchical deep learning architecture that exploits both local landmarks and general stent features by CNN backbones and achieves spatiotemporal associations using a GCN model. 3) We test the proposed method and several other state-of-the-art (SOTA) models on both public and private datasets, with hundreds of X-ray fluoroscopy videos, data of a scale that has not been reported before. The proposed method shows superior landmark detection, as well as frame-wise stent tracking performance.

\section{Related Work}
\label{sec:relatedwork}
\subsection{Digital Stent Enhancement}

The digital stent enhancement (DSE) algorithm typically follows a bottom-up design and can be generally divided in to 4 main steps: landmark detection, landmark tracking, frame registration, and enhanced stent display. 

First, the region of interest needs to be located from each frame. Due to the limited visibility of stent and large appearance variation between folded and expanded stents, it is challenging to extract the stent directly. Instead, potential landmarks such as the radiopaque balloon marker pairs at two ends of the stent or the guidewire in between is more commonly used to indicate the location of the stent. Throughout the X-ray image sequence, the stent location is constantly changing with cardiac and breathing motions. Based on the landmark candidates, the most promising track needs to be identified to associate the target stent across frames. Next, frames can be registered based on motion inferred by the landmark trajectory. The motion compensation is often performed by aligning tracked landmarks together using rigid registration~\cite{rigid1,rigid2} or elastic registration~\cite{elastic}. To enhance stent visualization, the stent layer is extracted while the background is suppressed~\cite{layer}.

\subsection{Balloon Marker Detection}
Two markers on the balloon used to deliver the stent are considered the most prominent feature of the stent structure due to the consistent ball shape and radio-opacity from high absorption coefficient. Various strategies are previously studied to achieve efficient balloon marker detection. 

Conventional image processing methods are applied, including match filtering or blob detection, to extract candidate markers from the X-ray image. Bismuth et al.\cite{compreh} proposed a method involving a priori knowledge and dark top hat preprocessing to detect potential markers from local minimum selection. Blob detectors~\cite{blob1,blob2,wang} locate markers by differentiating regions with unique characteristics from neighborhood, such as brightness or color. However, the extremely small size of balloon markers and common noises from the background, such as guidewire tips, Sharp bone edges and other marker-like structure, make those methods prone to a high false positive rate.

Learning-based methods are also proposed to incorporate more extended context information for better markers localization.
Lu et al.\cite{lu} used probabilistic boosting trees combining joint local context of each marker pair as classifiers to detect markers. Chabi et al.\cite{chabi} detected potential markers based on adaptive threshold and refined detections by excluding non-mask area using various machine learning classifiers, including k-nearest neighbor, naive Bayesian classifier, support vector machine and linear discriminant analysis. Vernikouskaya et al.\cite{unet} employed U-Net, a popular encoder-decoder like CNN designed specifically for medical images, to segment markers and catheter shafts during pulmonary vein isolation as binary masks. The maker segmentation performances from the above methods are still limited by the super imbalance between foreground and background areas. Moreover, all the candidate refinements only focus on considering more local context information at single frame, while the temporal correlation has never been exploited to enhance the classifiers.

In our work, balloon marker detection is considered as a heatmap regression task, which has shown superior performances in other landmark detection applications, such as face recognition~\cite{face}, human pose estimation~\cite{pose} and landmark detection in various medical images~\cite{medic1,medic2}. To obtain potential markers, we use U-Net as the heatmap regression model, which represents each landmark as a 2D Gaussian distribution for more accurate localization. 

\subsection{Graph Based Object Tracking}

Given the set of marker candidates across X-ray image sequence, either a priori motion information~\cite{compreh} or a heuristic temporal coherence analysis~\cite{lu}, which calculates consistency score between frames base on predefined criteria, is used to identify the most prominent landmark trajectory. 
Wang et al.\cite{wang} proposed a offline tracking algorithm as graph optimization problem, by constructing a trellis graph with all the potential marker pairs and then employed the Viterbi algorithm to search the optimal path across frame from the graph. Similar graph models are also applied to other general object tracking tasks~\cite{track1,track2,track3} as min-cost flow optimization problem. However, these static graph models will fail when the information contained by nodes or edges is not representative enough or outdated. Bras\'o et al.\cite{GNN} demonstrated superior results in multiple object tracking by constructing a dynamical graph of object detections and updating node and edge features using GCN. 

In this work, we first interpret the whole video into a graph, where the nodes are associated with encoded appearance features of potential stent from marker pair detections and edges are temporal coherency across frames. A graph neural network is trained as a node classification model to update both node and edge features via message passing. The stent tracking is achieve by learning both context and temporal information.

To our knowledge, the proposed CNN-GCN based DSE algorithm is the first deep learning model to achieve robust balloon marker tracking and 2D stent visual enhancement by incorporating both extended context and temporal information.

\section{Approach}
\label{sec:approach}
\begin{figure}[t!]
    \centering
    \includegraphics[width=\textwidth]{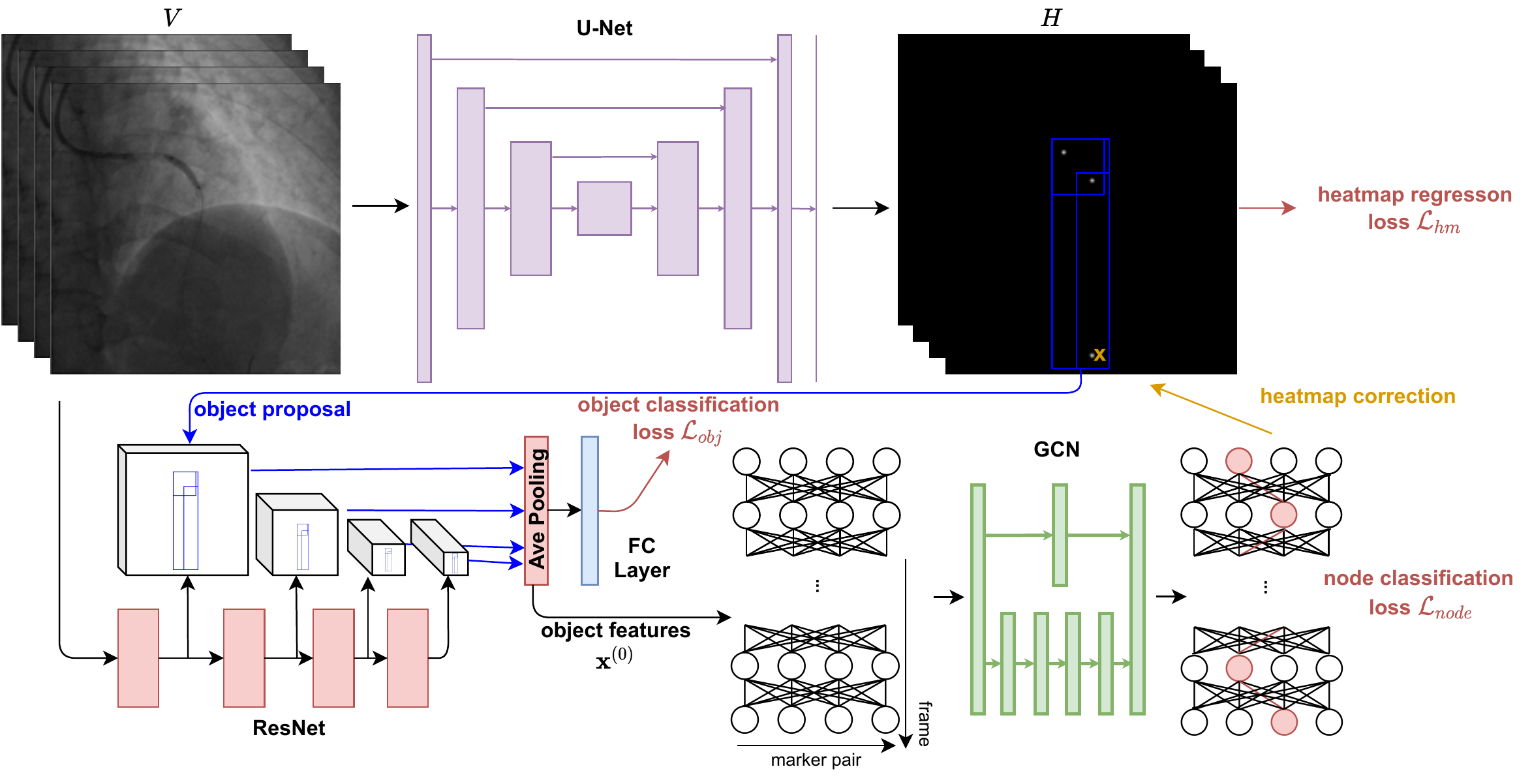}
    \caption{The proposed end-to-end deep learning framework for stent tracking.}
    \label{fig:method}
\end{figure}

In this work, we propose an effective end-to-end trainable framework for landmark based stent/balloon tracking (Fig.~\ref{fig:method}) with a hierarchical design: a U-Net based landmark detection module that generates a heatmap to localize marker candidates with local features, a ResNet based stent proposal and feature extraction module to extract global stent features in a larger context, and a GCN based stent tracking module to identify the real stent by temporally reasoning with stent features and marker locations.  

\subsection{Landmark Detection}
In landmark based stent/balloon detection, each candidate object is represented by a detected landmark pair: $\mathcal{O}_i = ( \mathcal{D}^{L}_{i1}, \mathcal{D}^{L}_{i2})$. The first step is to detect landmarks from each frame using a U-Net~\cite{unet}. In contrast to conventional object detection, the major challenge of landmark detection is the highly unbalanced foreground/background ratio, as landmarks are commonly tiny dots of few pixels compared to the frame size. Therefore, we treat landmark detection as a heatmap regression problem and pretrain the detector with smaller positive landmark patches, thus to increase fore-to-background ratio. The input video $\mathbf{V}\in\mathbb{R}^{T\times H\times W\times C}$ is fed into the landmark detector (U-Net) that generates heatmaps $\mathbf{H}\in\mathbb{R}^{T\times H\times W}$, where detected landmarks are represented as 2D Gaussian distributed points. From a predicted heatmap, peak points are extracted as landmark detections, represented as 2D coordinates and a confidence score: $\mathcal{D}^{L}_{i} = (x^{L}_{i},y^{L}_{i},s^{L}_{i})$. During training, an false negative regularization(\textbf{Sec~\ref{Training}}) is implemented to further enforce the detector to focus on landmarks.

With an ideal landmark detection, the target stents can be directly located by landmarks and tracked over time with simple temporal association. However, due to the lack of extended context information for perfect landmark localization, the landmark detector is inevitably limited by a high false positive rate which further hinders stent tracking. Hence, we proposed a delicate pipeline to simultaneously refine object detection and tracking.

\subsection{Stent Proposal \& Feature Extraction}
Given a set of landmark detections $\mathcal{D}^{L}_{t}$ at frame $(t)$, candidate objects can be formed by all possible combination of landmark pairs $\mathcal{O}_t = \{ (\mathcal{D}^{L}_{ti}, \mathcal{D}^{L}_{tj}) \mid \mathcal{D}^{L}_{ti}, \mathcal{D}^{L}_{tj} \in \mathcal{D}^{L}_{t} \}$, where  $\mathcal{D}^{L}_{ti}, \mathcal{D}^{L}_{tj}$ denote the $i$th and $j$th \underline{L}andmark \underline{D}etections in $\mathcal{D}^{L}_{t}$ at frame ($t$). As the landmark pair is always located at two ends of the corresponding object, We can assign a confidence score to the candidate object using the average of landmark confidence scores:
\begin{equation}
    S^\mathcal{O}_i = \frac{1}{2}({S}^{L}_{i1}+{S}^{L}_{i2}).
    \label{eq:ObjScore}
\end{equation}
We can generate a rectangular bounding box for the object based on the landmark locations, of which the center is the middle point of the landmark pair and side lengths are the distance between the landmarks along the corresponding axis. A ResNet is used to extract appearance features of candidate objects. The outputs of ResNet at multiple levels within the corresponding bounding boxes were averaged and stored into a D-dimension feature vector $\mathbf{x}^{(0)} \in \mathbb{R}^D$ for each candidate object, which are used later in GCN for temporal reasoning  (Fig.~\ref{fig:method}). In addition, to facilitate feature learning with a deep supervision, we feed the feature vector into a fully-connected layer and use a weighted cross-entropy loss ($\mathcal{L}_{obj}$ in Eq.~\ref{eq:total}) between the its outputs and labels indicating if the proposed bounding box contains the object of interests.

\subsection{Stent Tracking}
With the object candidates at every frame, we first construct an undirected graph $\mathcal{G} =(\mathcal{V},\mathcal{E})$ across all frames, where vertices $\mathcal{V}$ represent candidate objects proposed by detected landmark pairs and edges $\mathcal{E}$ are full connections of candidate objects between adjacent frames. Every object at frame ${t}$ is connected with all the candidate objects at frame $(t-1)$ and frame $(t+1)$. 

The attributes of vertices are the appearance feature vectors $\mathbf{x}^{(0)}$ extracted from the feature extractor. The edge weights in the initial graph are calculated as a weighted combination of object confidence scores and the spatial similarity by comparing sizes, rotations and locations of objects:
\begin{equation}
    \mathbf{w}_{i,j} = \frac{{S}^\mathcal{O}_i+{S}^\mathcal{O}_j}{2}( \alpha_1IoU(\mathcal{O}_i, \mathcal{O}_j) + \alpha_2AL(\mathcal{O}_i, \mathcal{O}_j)),
    \label{eq:EdgeWeight}
\end{equation}
where $\alpha_1$, $\alpha_2$ are weighting factors, $IoU(\cdot)$ is the IoU between object bounding boxes and $AL(\cdot)$ measures the objects similarity
by comparing angles and lengths of the landmark pair vector, defined as: 
\begin{equation}
    AL(\mathcal{O}_i, \mathcal{O}_j)) = \text{max}(0, \frac{|\vec{v}_i\cdot \vec{v}_j|}{| \vec{v}_i||\vec{v}_j|} - \frac{||\vec{v}_i|-|\vec{v}_j||}{ \sqrt{|\vec{v}_i||\vec{v}_j|}}).
    \label{eq:AL}
\end{equation}
Here, $\vec{v}_i$,$\vec{v}_j$ are the 2D vectors between landmark pairs of $\mathcal{O}_i$ and $\mathcal{O}_j$.

The initialized graph is a powerful representation of the whole video for object tracking, as the appearance features are embedded into vertices and spatiotemporal relations are embedded in the edge weights. To track objects over time, we perform node classification on the graph using a GCN, which identifies the tracked objects at different frames as positive nodes of a corresponding object class while false object detections and untracked objects are classified as negative nodes. 

The tracking model is a GCN with a full connected bypass (Fig.~\ref{fig:method}). The GCN branch consists of a weighted graph convolution layer (wGCL)~\cite{wGCL} and two edge convolution layers (ECLs)~\cite{edgeConv}. Weighted graph convolution layer with self-loop is defined as:
\begin{equation}
    \mathbf{x}_i = \mathbf{\Theta} \sum_{j \in \mathcal{N}(i) \cup\{ i \}} \frac{\mathbf{w}_{j,i}}{\hat{d}_i} \mathbf{x}^{(0)}_j,
    \label{eq:WeiGCN}
\end{equation}
with a normalization term $\hat{d}_i = 1 + \sum_{j \in \mathcal{N}(i)} \mathbf{w}_{j,i}$, where
$\mathbf{w}_{j,i}$ denotes the edge weight between node $j$ and node $i$.

Within an edge convolution layer, the edge features are first updated by a FC layer with the features of corresponding vertices pairs connected with each edge:
\begin{equation}
    \mathbf{e}_{i,j} = h_{\mathbf{\Theta}}(\mathbf{x}_i, \mathbf{x}_j),
  \label{eq:EdgeFeat}
\end{equation}
where $h_{\mathbf{\Theta}}$ is a nonlinear function with learnable parameters $\mathbf{\Theta}$. Then, ECL updates node features by the summation of updated edge features associated with all the edges emanating from each vertex:
\begin{equation}
    \mathbf{x}^{EC}_i = \sum_{j \in \mathcal{N}(i)} \mathbf{e}_{i,j}.
  \label{eq:EdgeConv}
\end{equation}

The GCN branch effectively updates features of candidate objects by most similar objects from adjacent frames. Moreover, a sequence of convolution layers enables information propagation from even further frames. However, node features solely updated from the GCN are susceptible to noisy neighborhood. For example, if the target object is missed by the upstream detection at a certain frame, such errors would propagate to nearby frames and thus worsen general tracking performance. Therefore, we add a simple parallel FC bypass to the GCN branch. In the FC bypass, all the node features are updated independently without influence from connected nodes:
\begin{equation}
    \mathbf{x}^{FC}_i = h_{\mathbf{\Theta}}(\mathbf{x}^{(0)}_i).
  \label{eq:FC}
\end{equation}

In the last layer, node features from the GCN branch $\mathbf{x}^{EC}$ are enhance by the FC bypass outputs $\mathbf{x}^{FC}$ for robust object tracking.

{\bf Heatmap correction} GCN results are then used to correct the heatmaps generated in landmark detection. Specifically, we multiply heatmap values in the a $w \times w$ window centered around a detected landmark with the maximum probability of the graph nodes containing the marker. In this way, the landmark detector can ignore the false positives that can be easily rejected by the GCN model and increase detection sensitivity. 

\subsection{Training}
\label{Training}
The landmark detector was trained as a heatmap regression model. Since landmark detection results are used for object proposal, feature extraction, and graph construction, missed landmark would cause irreversible corruption to tracking as the missed object cannot be recovered, while false positives can be filtered out during object proposal or node classification. We used a modified cost term $\mathcal{L}_{hm}$ to ensure fewer false negatives, defined as:

\begin{equation}
    \mathcal{L}_{hm} = \frac{\lambda_1}{N} \sum_{i=0}^{N}(y_i - {\hat{y}}_i)^2 + \frac{\lambda_2}{N} \sum_{i=0}^{N}(ReLU(y_i - {\hat{y}}_i))^2,
  \label{eq:HMcost}
\end{equation}
where $\lambda_1$, $\lambda_2$ are weighting factors, $y_i$, ${\hat{y}}_i$ are pixel intensities from ground truth and predicted heatmap (corrected by GCN outputs), respectively.

The feature extractor and GCN tracking model are trained as classification problems. We use weighted cross entropy as the cost function to handle the unbalanced labels (most object candidates are negative), which is defined as $-\sum^{C}_{i=1}{w_ip_i\log(\hat{p}_i)}$, where $p$, $\hat{p}$ denote the ground truth and predicted object/node class, respectively, and $w_i$ is the predefined weight for class $i$. 

Taken together, the total loss for end-to-end training is
\begin{equation}
    \mathcal{L} = \mathcal{L}_{hm} + \alpha\mathcal{L}_{obj} + \beta\mathcal{L}_{node},
    \label{eq:total}
\end{equation}
where $\mathcal{L}_{obj}$ and $\mathcal{L}_{node}$ are weighted cross entropy losses for object classification and node classification respectively.

\section{Experiments}
\label{sec:experiments}
\subsection{Datasets}
 Our in-house stent dataset consists of 4,480 videos (128,029 frames of 512 $\times$ 512 8-bit frames) acquired during PCI procedures. The data acquisition was approved by Institutional Review Boards. For in-house videos, the landmarks are radiopaque balloon marker pairs located at two ends of the stent (Fig.~\ref{fig:teaser}). There are 114,352 marker pairs in the dataset, which were manually annotated by trained experts. The dataset was split into training, validation, and testing set with a 8:1:1 ratio, which resulted in 3584 videos (103892 frames), 448 videos (12990 frames), and 448 videos (11147 frames) respectively.
 
 In addition, to verify generalization of our method, we included a public dataset in our experiment. The transcatheter aortic valve implantation (TAVI) dataset is a public intraoperative aortography X-ray imaging dataset including 35 videos of $1000\times1000$ pixels 8-bit frames. The original dataset consisted of 11 keypoint annotations including 4 anatomical landmarks, 4 catheter landmarks and 3 additional background landmarks. TAVI is different from PCI procedure but it contains landmark pairs: Catheter Proximal (CP) and Catheter Tip (CT) whose constellations are similar to the stents. We excluded irrelevant landmarks resulting the final TAVI dataset contains 2,652 frames from 26 videos. The TAVI dataset was randomly divided into a training set with 2,027 images from 18 videos and a test set with 625 images from 8 videos. We ran K-fold (k=5) cross-validation for all models on both private dataset and the TAVI dataset, and the detailed results are reported in the Supplementary Materials. 
\subsection{Comparative Models}
To demonstrate the efficacy of our algorithm, we compared it with several SOTA models on both datasets. First, we selected two coordinate regression models, ResNet V2~\cite{resnetv2} and MobileNet V2~\cite{MobileNet}. 
Such regression models detect landmarks in each frame by predicting the landmark center coordinates,  which have shown superior performance regressing TAVI catheter landmarks in \cite{TAVI}. Moreover, we include a center based multi-object tracking (MOT) model, CenterTrack~\cite{centertrack}, with two most powerful backbones: DLA-34~\cite{DLA} and MobileNet V2. CenterTrack detects objects as heatmap regression of centers and simultaneously tracks objects over time by predicting the translations of center points between adjacent frames. CenterTrack has demonstrated extraordinary performance on various MOT benchmark datasets.

\subsection{Evaluation Metrics}
As our final goal is to enhance the stents by aligning landmark points across frames, detection success rate and localization accuracy are the most important factors to ensure high-quality enhancement. To compare the landmark prediction performance, we use the following detection and localization metrics for evaluation. For detection performance, we used $Precision$, $Recall$, $F_1$ and $Accuracy$. Landmark locations extracted from heatmaps were paired with the closest ground truth(GT) greedily. A stent prediction was matched if distances of its both landmarks to paired GT were smaller than 5pxs(in-house) or 15pxs(TAVI).
\begin{gather*}
Precision =\frac{TP}{TP+FP}; \quad Recall =\frac{TP}{TP+FN} \\
F_1 =\frac{2\cdot TP}{2\cdot TP+FN+FP}; \quad Accuracy=\frac{TP+TN}{TP+FN+TN+FP}
\end{gather*}

On the successfully detected landmarks, we also evaluated landmark localization accuracy using pixel-wise MAE and RMSE:
$$\textit{MAE} = \frac{1}{N}\sum_{i=1}^{N}|p_i-\hat{p}_i|; \quad \textit{RMSE} = \sqrt{\frac{1}{N}\sum_{i=1}^{N}(p_i-\hat{p}_i)^2},$$
where $p_i$, $\hat{p}_i$ denote predicted and ground truth landmark coordinates.

\subsection{Implementation Details}
All deep learning models were implemented with PyTorch and run on NVIDIA V100. For the proposed method, the marker detection module was pre-trained on 128 $\times$ 128 image patches, and then the whole model was trained on 10-frame video clips. We used Adam optimizer with a learning rate of 1e-5. For the coordinates regression models, we follow the multi-task learning schemes provided by Danilov et al.~\cite{TAVI}, using Binary Cross Entropy for the classification branch and Log-Cosh loss for the regression branch, optimized with Adam with a learning rate of 1e-5. Similarly for CenterTrack, we trained the models based on the configuration described in the original publication. The major modification we made was to remove the branch of object bounding box size regression since we do not need the landmark size estimation for our task. We used the focal loss for heatmap regression and L1 loss for offset regression, optimized with Adam with learning rate 1.25e-4. Please see the supplementary material for more details on hyperparameters.

\section{Results and Discussion}

\label{sec:results}
\subsection{Main Results}
\setlength{\tabcolsep}{4pt}
\begin{table}[t]
\begin{center}
\caption{Evaluations on \textbf{In-house Dataset}. CR means coordinate regression model, and CT means CenterNet. $\uparrow$ indicates that higher is better, $\downarrow$ indicates that lower is better.}
\label{table:Inhouse}
\begin{tabular}{cc|cccc|cc}
\hline
\multicolumn{2}{c|}{Model}  &\multicolumn{4}{c|}{Detection}  &\multicolumn{2}{c}{Localization} \\
Type & Backbone& Precision$\uparrow$ & Recall$\uparrow$ & F1$\uparrow$ & Accuracy$\uparrow$ & MAE$\downarrow$ & RMSE$\downarrow$ \\
\hline
\multirow{2}{*}{CR} & MobileNetV2 &0.620 &0.557 & 0.587 & 0.415 & 1.172 & 1.283\\
&  ResNetV2 &0.618 &0.604 &0.611 &0.440 &1.064 &1.125\\
\hline
\multirow{2}{*}{CT} & MobileNetV2 &0.485& 0.932& 0.638& 0.469& 0.455& 0.837\\
& DLA34 &0.591&	\textbf{0.936} & 0.725& 0.568& \textbf{0.398}& \textbf{0.748}\\
\hline
\multicolumn{2}{c|}{Ours}  &\textbf{0.907} &0.908 & \textbf{0.908} & \textbf{0.831} & 0.597 & 0.963\\
\hline
\end{tabular}
\end{center}
\caption{Evaluations on \textbf{TAVI Dataset}.}
\label{table:TAVI}
\begin{tabular}{cc|cccc|cc}
\hline
\multicolumn{2}{c|}{Model}  &\multicolumn{4}{c|}{Detection}  &\multicolumn{2}{c}{Localization} \\
Type & Backbone& Precision$\uparrow$ & Recall$\uparrow$ & F1$\uparrow$ & Accuracy$\uparrow$ & MAE$\downarrow$ & RMSE$\downarrow$\\
\hline
\multirow{2}{*}{CR} & MobileNetV2&0.839& 0.735& 0.784& 0.644& 12.904& 14.129\\
& ResNetV2 &0.857& 0.846& 0.851& 0.741& 11.571& 12.490\\
\hline
\multirow{2}{*}{CT} & MobileNetV2 &0.785& \textbf{0.961}& 0.864&	0.761& \textbf{5.100}& \textbf{6.159}\\
& DLA34 &0.868& 0.930& 0.898& 0.815& 5.418& 6.357\\
\hline
\multicolumn{2}{c|}{Ours}  &\textbf{0.918}& 0.957& \textbf{0.938}& \textbf{0.882}& 5.975& 6.831\\
\hline
\end{tabular}
\end{table}

Table~\ref{table:Inhouse} and Table~\ref{table:TAVI} list the results of proposed model and baseline models on the in-house dataset and the public TAVI dataset~\footnote{Example of stent tracking and enhancement comparisons are included in the Supplementary Materials.}. The results are consistent on both datasets. In terms of detection, our framework significantly outperforms the prior state of the art on both datasets. Firstly, tracking models generally excels pure detection models as the additional temporal information is helpful to enhance landmark detection. Another major limitation of the coordinate regression models is that the number of detections is always fixed. Therefore, some targeted landmarks can be easily overwhelmed by strong background noises in coordinate regressor, resulting in a common trend of lower $recall$. On the contrary, heatmap regression models have the flexibility to predict more possible landmarks as long as the desired features are identified from the image. This would help achieve higher recall but also resulting in a large number of false positive landmark detections,  indicated as the worse CenterTrack $precision$ values compared to coordinate regressors. To solve this issue, in our framework design, we introduced both additional spatial information and temporal information to refine the noisy preliminary detections. 

As tracking models, the proposed model shows a remarkable detection margin over CenterTrack. The results demonstrate the effectiveness of our two major innovation in the framework: stent proposal and GCN-based tracking. Instead of tracking multiple landmarks as individual points as in CenterTrack, our model enhanced the isolated detections by introducing stent proposal stage. The feature of possible stents patches between candidate landmark pairs enables the model to enhance landmarks learning and context relationship between landmark pairs. Moreover, as a pure local tracking model, landmarks association of CenterTrack is limited to adjacent frames, where the landmark association is simply learnt as spatial displacements. The information propagation of our multi-layer GCN model enables stent feature nodes in graph, with the combination of both spatial and appearance features, to interact with other nodes in longer time range. We will further prove the effectiveness of our designs in the following ablation studies. 

To compare two datasets, our in-house dataset includes more videos with more complicated background compared to the TAVI dataset, which makes the stent tracking a more challenge task for all models, with more likely false positive detections. From the results, we could observe the notable declines in in-house dataset detection metrics, especially in $precision$, for each model compared to TAVI's. However, the $precision$ of our framework only dropped from $0.918$ to $0.907$, which indicates that our framework is more robust to suppress false positives and maintain high accuracy detection in more complex PCI scenes.
This robustness advantage is a good reflection of our hierarchical landmark and stent feature extraction efficacy at the stent proposal stage. Even though more complicated background would cause confusion to individual landmark detection, our model can still successfully target the desired landmark pairs by identifying the prominent stent feature in between. 

As for localization evaluation, the MAE and MSE values we got from TAVI is about 10 times larger than our own dataset. This is because, firstly, the original frame size is significantly larger than our data; secondly, the landmarks in TAVI, CP and CD, are also about 10 times larger than our landmarks. For example, the CD landmarks are more of a dim blob rather than a single opaque point as in our dataset. To compare the results among models, heatmap regression models perform generally better than coordinate regression models, as the numerical regression still remain an quite arbitrary task for CNN model and localization accuracy would be diminished during the single downsampling path in regressors. CenterTrack achieves the lowest MAE and MSE errors on both datasets. The high accurate localization of CenterTrack is realized by both sophisticated model architecture and the additional two-channel output branch specifically designed for localization offsets regression. The cost for the localization accuracy improvement is computational complexity and time (CenterTrack-DLA34: $10.1$ FPS). As for our model, the localization solely depends on the heatmap regression from the faster U-Net ($21.7$ FPS). We have not tried to add any localization refinement design for the framework efficiency. On our dataset, the mean MAE value at $0.597$ is already sufficient for our final stent enhancement task. Further localization refinement with the sacrifice of time would not cause any noticeable improvements. However, we want to note that our framework is very flexible that the current U-Net backbone can be replaced by any other delicate heatmap regression models, if a more accurate localization is necessary for an application.

\subsection{Ablation Studies}

The proposed end-to-end learning framework consists of three major modules: heatmap regression based landmark detection (U-Net), landmark-conditioned stent proposal and feature extraction (ResNet), and stent tracking GCN. To demonstrate the benefits of our framework design, we ablated the main components of stent proposal ResNet and stent tracking GCN, as well as the end-to-end learning regime. 

\textbf{Detection only} directly utilizes the U-Net to detect individual landmarks at each frame. This model only needs separate frames as inputs and individual landmark locations as supervision. 

\textbf{Detection refinement} improves landmark detections in the heatmap prediction from the U-Net backbone and filters false positives by incorporating additional spatial information of the stent between candidate landmarks using a ResNet patch classifier. This two-step model also only requires single frame inputs and no temporal association is used.

\textbf{Separate Learning} includes all three proposed major modules (landmark detection, stent proposal and stent tracking). Instead of simply filtering out false stent patches by the CNN model, this model first uses the convolution layers from well-trained patch classification ResNet to extract feature vectors from candidate stent patches. Then, the features are reconstructed into stent graph and fed into the GCN model for stent tracking over frames. The final model outputs would be the tracked stents at each frame of the input video. However, this three-step approach is achieved by training each model independently: U-Net for landmark detection, ResNet for stents patch classification and GCN for stent tracking.  
\setlength{\tabcolsep}{4pt}
\begin{table}[t]
\caption{Ablation study on \textbf{In-house Dataset}.}
\label{table:ablation}
\begin{tabular}{c|cccc|cc}
\hline
\multirow{2}{*}{Model}  &\multicolumn{4}{c|}{Detection}  &\multicolumn{2}{c}{Localization} \\
\cline{2-7}
& Precision$\uparrow$ & Recall$\uparrow$ & F1$\uparrow$ & Accuracy$\uparrow$ & MAE$\downarrow$ & RMSE$\downarrow$\\
\hline
Detection only & 0.551 & \textbf{0.918} & 0.688 & 0.525 & \textbf{0.596} & 0.998\\
Detection refinement & \textbf{0.927} & 0.532 & 0.677 & 0.511 & 0.598 & \textbf{0.946}\\
Separate learning &0.893 &0.872 & 0.883 & 0.790 & \textbf{0.596} & 0.955\\
\hline
Ours  & 0.907 &0.908 & \textbf{0.908} & \textbf{0.831} & 0.597 & 0.963\\
\hline
\end{tabular}
\end{table}

The ablation study results are shown in Table~\ref{table:ablation}. Our full end-to-end model performs significantly better than the baselines in the detection task. The standalone U-Net yields a very high false positive rate ($44.9\%$) as it is difficult for this model to learn meaningful features to differentiate small landmarks from dark spot noises in the background.

In the detection refinement results, the stent patch classifier significantly reduced the false positives from U-Net predictions, as $precision$ surged to $92.7\%$. However, simply applying the patch classifier to the U-Net outputs would also filter out true stent patches with weaker patch features, resulting in a large drop in $recall$. The above results indicate a trade-off between $precision$ and $recall$ while applying spatial information based models.

The results of separate learning demonstrate that incorporating GCN temporal stent tracking improved recall and maintained the high precision from detection refinement, resulting in a boost in overall detection accuracy. False negatives are effectively suppressed by the information propagation mechanism in GCN, which helps to enhance the feature of weak but true stent nodes with nearby strong stent nodes in both space and time. 

Compared with the separate three-stage learning model, our proposed end-to-end model achieved further improvements in all detection evaluation metrics and reached a better balance between $precision$ and $recall$. Although different components of our framework have their specific tasks along the detection and tracking process, the end-to-end learning brings extra benefits, especially by optimizing the data flow between modules. For example, the back-propagated gradients from GCN can also guide the convolutional layers at stent proposal to extract better patch features that would be fed into GCN.

In regard to localization accuracy, all baseline models and the final model show similar performance, as we used the same U-Net backbone for all experiments. The MAE and RMSE values fluctuate within $0.002$ and $0.052$, which we believe are only from experimental uncertainty and would not have a sensible influence on the final stent enhancement task. For many multi-task learning models on limited data, there is conventionally a trade-off between excellency on specific metrics and good overall performance. The results suggest that the complicated multi-task learning of our end-to-end model would both maintain high localization accuracy and improve detection.

\section{Conclusion}
\label{sec:conclusion}
In this work, we proposed a novel end-to-end CNN-GCN framework for stent landmarks detection and tracking. The model includes three major modules: (1) U-Net based heatmap regression for landmark candidate detection, (2) a ResNet for landmark-conditioned stent proposal and feature extraction, and (3) residual-GCN based stent tracking. We compared the proposed model with SOTA coordinate regression models and multi-object tracking models. Our experiments demonstrated that the proposed model remarkably outperformed previous SOTA models in stent detection. We further discussed the flexibility of the proposed framework to accommodate new heatmap regression backbones to overcome the current localization limitations. The ablation experiments showed the benefits of our novel designs in stent proposal ResNet, stent tracking GCN, and end-to-end learning scheme.

\clearpage

%
%
\bibliographystyle{splncs04}
\bibliography{refer}

\title{Supplementary Material} 


\author{}
\institute{}

\maketitle
\section{Hyperparameters}
\textbf{Neural network architecture} We adapted the default architecture in \cite{ronneberger2015u} for our U-Net marker detection module. The difference was that we only used three downsampling operations to prevent overfitting. We used ResNet-18 \cite{He_2016_CVPR} for the object feature extraction module. For the GCN tracking module, we used the PyTorch Geometric library \footnote{https://pytorch-geometric.readthedocs.io/en/latest/}. The GCN branch contains a \textit{GCNConv} layer (1024 input channels and 256 output channels), a \textit{EdgeConv} layer (512 input channels and 128 output channels), and a \textit{EdgeConv} layer (256 input channels and 64 output channels), with a \textit{ReLU} layer after each layer. The FC branch has one fully-connected layer with 1024 input channels and 256 output channels. Then outputs of GCN branch and FC branch were combined and passed to a fully-connected layer with 320 input channels and 1 output channel.

\textbf{Loss function} In Eq. 8, $\lambda_1 = 1, \lambda_2 = 2$. In Eq. 9, $\alpha = 1, \beta = 2$. 
\section{Results}
\begin{figure}
    \centering
    \includegraphics[width=\textwidth]{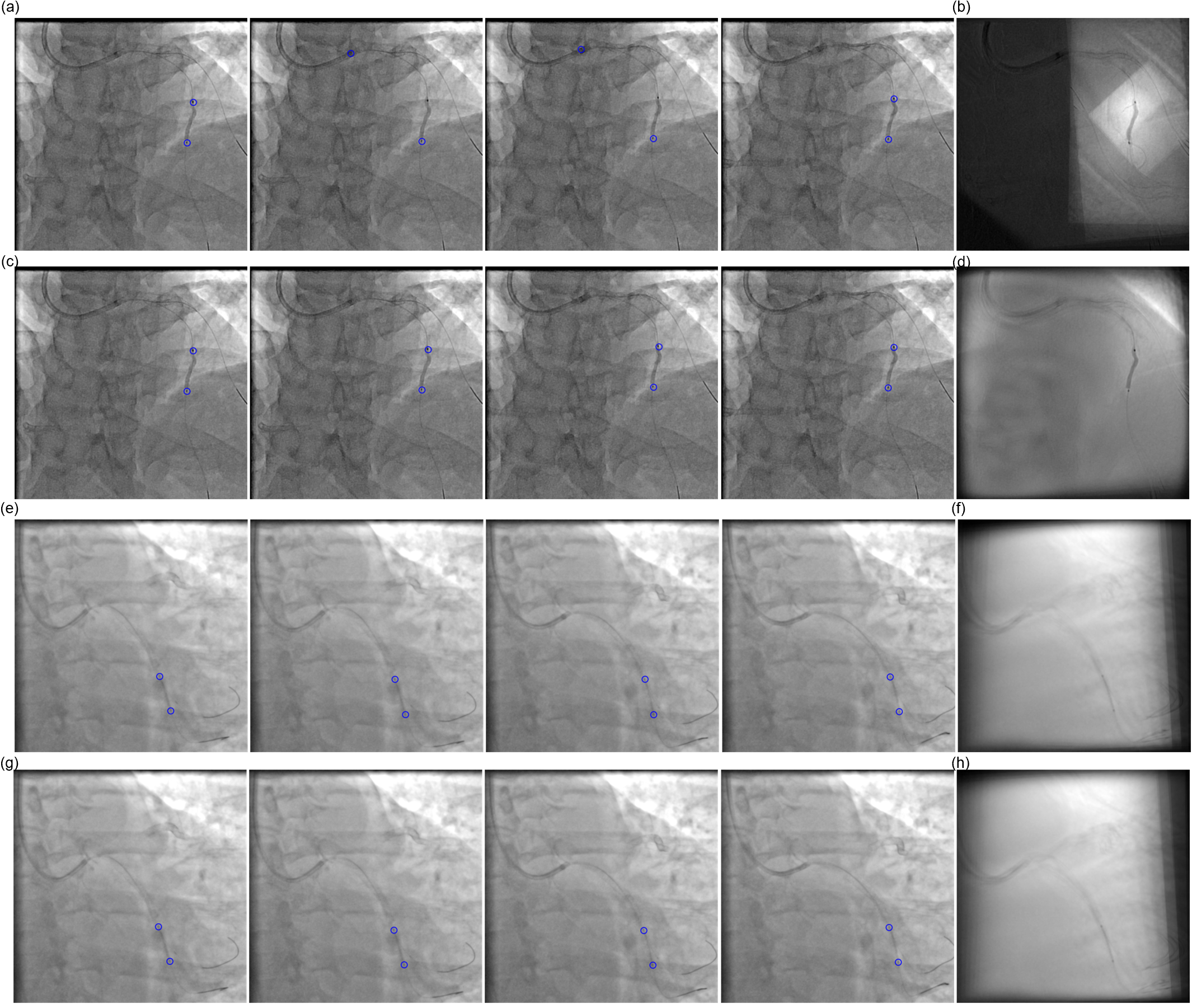}
    \caption{Example tracking results (4 frames) from CenterTrack (a,e) and the proposed method (c,g), and the corresponding stent enhancement results (b,f) and (d,h). 7 frames were used for enhancement in all cases.}
    \label{fig:results}
\end{figure}

We used a fixed threshold cutoff (0.6) for identifying positive predictions (i.e. markers with probability greater than 0.6 were selected) to calculate the evaluation metrics shown in Table 1, 2, and~\ref{table:cv}. Here, we also demonstrate the performance of CenterTrack and our method (Table~\ref{table:Inhouse-top2} and Table~\ref{table:TAVI-top2}) when two markers with the highest probabilities in each frame were identified as markers detected, since it is the most straightforward way to identify a single stent based on the outputs of neural networks. Fig.~\ref{fig:results} shows examples of tracking results from CenterTrack and the proposed method with this top-2 selection criterion, and the stent enhancement results based on the tracking results. It can observed that false positives dramatically affect enhancement results (Fig.~\ref{fig:results}b) and our method has a high precision score (Table~\ref{table:Inhouse-top2} and Table~\ref{table:TAVI-top2}), which demonstrates its robustness in clinical applications. Lower two rows of Fig.~\ref{fig:results} show results of CenterTrack and the proposed method with representative MAEs (0.382 and 0.511) respectively. It can be observed that even though the MAE of our method is worse than that of CenterTrack, the enhancement results do not show much difference.

\begin{table}
\begin{center}
\caption{Evaluations on \textbf{In-house Dataset} (top 2). CR means coordinate regression model, and CT means CenterNet. $\uparrow$ indicates that higher is better, $\downarrow$ indicates that lower is better.}
\label{table:Inhouse-top2}
\begin{tabular}{cc|cccc|cc}
\hline
\multicolumn{2}{c|}{Model}  &\multicolumn{4}{c|}{Detection}  &\multicolumn{2}{c}{Localization} \\
Type & Backbone& Precision$\uparrow$ & Recall$\uparrow$ & F1$\uparrow$ & Accuracy$\uparrow$ & MAE$\downarrow$ & RMSE$\downarrow$ \\
\hline
\multirow{2}{*}{CT} & MobileNetV2 &0.752& 0.803& 0.777 & 0.635 & 0.443 & 0.827 \\
& DLA34 &0.813&	0.805 & 0.809 & 0.679 & \textbf{0.391}& \textbf{0.742}\\
\hline
\multicolumn{2}{c|}{Ours}  &\textbf{0.979} & \textbf{0.882} & \textbf{0.928} & \textbf{0.866} & 0.502 & 0.891\\
\hline
\end{tabular}
\end{center}
\caption{Evaluations on \textbf{TAVI Dataset} (top 2).}
\begin{center}
\label{table:TAVI-top2}
\begin{tabular}{cc|cccc|cc}
\hline
\multicolumn{2}{c|}{Model}  &\multicolumn{4}{c|}{Detection}  &\multicolumn{2}{c}{Localization} \\
Type & Backbone& Precision$\uparrow$ & Recall$\uparrow$ & F1$\uparrow$ & Accuracy$\uparrow$ & MAE$\downarrow$ & RMSE$\downarrow$\\
\hline
\multirow{2}{*}{CT} & MobileNetV2 &0.919 & 0.905 & 0.907&	0.831& \textbf{5.172} & \textbf{6.054}\\
& DLA34 &0.927& 0.896& 0.911& 0.837& 5.415& 6.150\\
\hline
\multicolumn{2}{c|}{Ours}  &\textbf{0.986} & \textbf{0.915} & \textbf{0.949}& \textbf{0.903}& 5.966& 6.705\\
\hline
\end{tabular}
\end{center}
\end{table}

\begin{table}
\begin{center}
\caption{Evaluations on \textbf{TAVI Dataset} with cross validation. CR means coordinate regression model, and CT means CenterNet. All the reported values are mean values from 5-fold cross validation. $\uparrow$ indicates that higher is better, $\downarrow$ indicates that lower is better.}
\label{table:cv}
\begin{tabular}{cc|cccc|cc}
\hline
\multicolumn{2}{c|}{Model}  &\multicolumn{4}{c|}{Detection}  &\multicolumn{2}{c}{Localization} \\
Type & Backbone& Precision$\uparrow$ & Recall$\uparrow$ & F1$\uparrow$ & Accuracy$\uparrow$ & MAE$\downarrow$ & RMSE$\downarrow$ \\
\hline
\multirow{2}{*}{CR} & MobileNetV2&0.851& 0.741& 0.7892& 0.656& 13.44& 14.56\\
& ResNetV2 &0.861& 0.799& 0.829& 0.709& 12.145& 13.168\\
\hline
\multirow{2}{*}{CT} & MobileNetV2 &0.751& \textbf{0.934} & 0.832& 0.713& \textbf{5.276}& \textbf{6.248}\\
& DLA34 &0.819&	0.927 & 0.869 & 0.768& 5.362 & 6.490\\
\hline
\multicolumn{2}{c|}{Ours}  &\textbf{0.913} &0.902 & \textbf{0.901} & \textbf{0.820} & 5.802 & 6.524\\
\hline
\end{tabular}
\end{center}
\end{table}
\end{document}